\documentclass[review]{elsarticle}

\usepackage{lineno,hyperref}
\usepackage{booktabs}
\usepackage{graphicx}
\usepackage{adjustbox}
\usepackage[dvipsnames]{xcolor}
\usepackage{multirow}
\usepackage{amsmath}
\usepackage{amssymb}

\usepackage[ruled,vlined,linesnumbered,commentsnumbered]{algorithm2e}
\usepackage{bm}
\usepackage{float}
\usepackage{longtable}
\restylefloat{table}

\modulolinenumbers[5]

\journal{Journal of \LaTeX\ Templates}

\begin{document}

\begin{frontmatter}

\title{Automated Design of Salient Object Detection Algorithms with Brain Programming}

\author[1]{Gustavo Olague}
\cortext[mycorrespondingauthor]{Corresponding author}
\ead{olague@cicese.mx}
\author[1]{Jose Armando Menendez-Clavijo}
\author[4]{Matthieu Olague}
\author[3]{Arturo Ocampo}
\author[2]{Gerardo Ibarra-Vazquez}
\author[5]{Rocio Ochoa}
\author[1]{Roberto Pineda}

\address[1]{EvoVisi\'{o}n Laboratory, CICESE Research Center. Carretera Ensenada-Tijuana 3918, Zona Playitas, 22860, Ensenada, B.C. M\'{e}xico}
\address[2]{Universidad Autónoma de San Luis Potos\'{\i}, Facultad de Ingenier\'{\i}a. Dr. Manuel Nava 8, Col. Zona Universitaria Poniente, 78290, San Luis Potos\'{\i}, S.L.P., M\'{e}xico}
\address[3]{Universidad Nacional Aut\'{o}noma de M\'{e}xico, Av Hacienda de Rancho Seco S/N, Impulsora Popular Avicola, 57130 Nezahualc\'{o}yotl, M\'{e}xico}
\address[4]{Universidad Anahuac Quer\'{e}taro, Calle Circuito Universidades I,
Kil\'{o}metro 7, Fracci\'{o}n 2, El Marqu\'{e}s, Quer\'{e}taro. C.P.76246, M\'{e}xico}
\address[5]{Universidad Aut\'{o}noma de Tlaxcala, Facultad de Ciencias B\'{a}sicas Ingenier\'{\i}a y Tecnolog\'{\i}a, Carretera Apizaquito S/N, San Luis Apizaquito, C.P. 90401, Apizaco, Tlaxcala, M\'{e}xico}

%
%
%
%

\begin{abstract}
Despite recent improvements in computer vision, artificial visual systems' design is still daunting since an explanation of visual computing algorithms remains elusive. Salient object detection is one problem that is still open due to the difficulty of understanding the brain's inner workings. Progress on this research area follows the traditional path of hand-made designs using neuroscience knowledge. In recent years two different approaches based on genetic programming appear to enhance their technique. One follows the idea of combining previous hand-made methods through genetic programming and fuzzy logic. The other approach consists of improving the inner computational structures of basic hand-made models through artificial evolution. This research work proposes expanding the artificial dorsal stream using a recent proposal to solve salient object detection problems. This approach uses the benefits of the two main aspects of this research area: fixation prediction and detection of salient objects. We decided to apply the fusion of visual saliency and image segmentation algorithms as a template. The proposed methodology discovers several critical structures in the template through artificial evolution. We present results on a benchmark designed by experts with outstanding results in comparison with the state-of-the-art.
\end{abstract}


\end{frontmatter}


\section{Introduction}

Saliency is a property found in the animal kingdom whose purpose is to select the most prominent region on the field of view. Elucidating the mechanism of human attention including the learning of bottom-up and top-down processes is of paramount importance for scientists working at the intersection of neuroscience, computer science, and psychology. Giving a robot/machine this ability will allow it to choose/differentiate the most relevant information. Learning the algorithm for detecting and segmenting salient objects from natural scenes has attracted great interest in computer vision and recently by people working with genetic programming \cite{Dozal2014,Contreras-Cruz2019}. While many models and applications have emerged, a deep understanding of the inner workings remains lacking. This work develops over a recent methodology that attempts to design brain-inspired models of the visual system, including dorsal and ventral streams \cite{Clemente2012,avc}. The dorsal stream is known as the ``where'' or ``how'' stream. This pathway is where the guidance of actions and recognizing objects' location in space is involved and where visual attention occurs. The ventral stream is known as the ``what'' stream. This pathway is mainly associated with object recognition and shape representation tasks. This work deals with the optimization/improvement of an existing algorithm (modeling the dorsal stream) and allows evolution to improve this initial template method. The idea is to leverage the human designer with the whole dorsal stream design's responsibilities by focusing on the high-level concepts while leaving the computer (genetic programming-GP) the laborious chore of providing optimal variations to the template. Therefore, the human designer is engaged in the more creative process of defining a family of algorithms \cite{ROJASQUINTERO2021103834}. 

\begin{figure*}
    \centering
   \includegraphics[width=\textwidth]{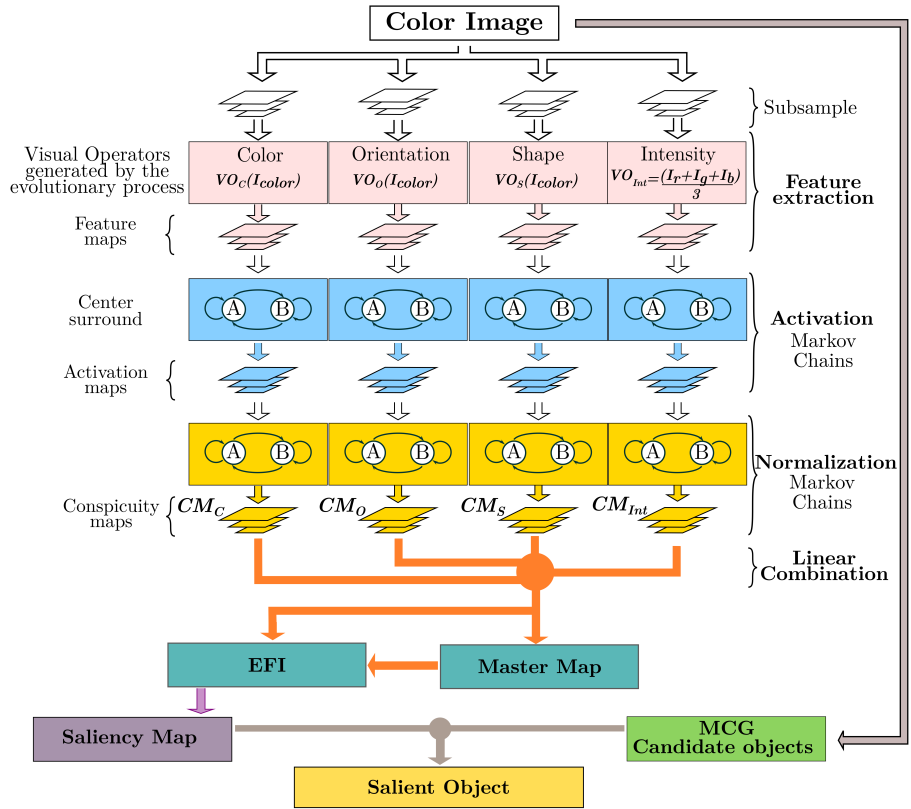}
    \caption{Brain programming implementation of the dorsal stream using the combination of visual saliency and image segmentation algorithms. We propose to discover a set of visual operators ($VOs$) and the evolutionary feature integration (EFI) within the template through artificial evolution. The whole design makes a design balance between the human designer and the computer.}
    \label{fig:evoGBVS}
\end{figure*}

Figure \ref{fig:evoGBVS} shows the template's implementation (individual representation) that emulates an artificial dorsal stream (ADS). As we can observe, the whole algorithm represents a complex process based on two models. A neurophysiological model called the two pathway cortical model--the two streams hypothesis--and a psychological model called feature integration theory \cite{Treisman1980}. This last theory states that human beings perform visual attention in two stages. The first is called the preattentive stage, where visual information is processed in parallel over different feature dimensions that compose the scene: color, orientation, shape, intensity. The second stage, called focal attention, integrates the extracted features from the previous stage to highlight the scene's region (salient object). Hence, the image is decomposed into several dimensions to obtain a set of conspicuity maps, which are then integrated--through a function known as evolved feature integration (EFI)--into a single map called the saliency map. Brain programming (BP) is based on the most popular theory of feature integration for the dorsal stream and the hierarchical representation of multiple layers as in the ventral stream \cite{Khan2021}. Note that the template's design can be adapted according to the visual task. In this work, we focus on designing an artificial dorsal stream. Moreover, BP replaces the data-driven models with a function-driven paradigm. In the function-driven process, a set of visual operators ($VOs$) is fused by synthesis to describe the image's properties to tackle object location and recognition tasks.

This paper is organized as follows. First, we outline the related work briefly to highlight the research direction. Next, we detail the construction of the ADS template using an adaptation of Graph-based Visual Saliency (GBVS) combined with the Multiscale Combinatorial Grouping (MCG) to an evolutionary machine learning algorithm. Then, we present the results of the evolutionary algorithm to illustrate the benefits of the new proposal. Finally, we finish the article with our conclusions and future work on the automate design of brain models.

\section{Related Work}

For a learning algorithm design technique to be well received, it needs to solve several analysis levels. A major critique of deep learning is the opacity. Scientists depend on complex computational systems that are often ineliminably opaque, to the detriment of our ability to give scientific explanations and detect artifacts. Here we follow a strategy for increasing transparency based on three levels of explanation about what vision is and how it works and why we still lack a general model, solution, or explanation for artificial vision \cite{Creel2020}. The idea follows a goal-oriented framework where learning is studied as an optimization process \cite{Olague2016}. The first is theoretical transparency or knowledge of the visual information processing whose design is the computation goal. The second is algorithmic transparency or knowledge of visual processing coding. Finally, the third level is execution transparency or knowledge of implementing the program considering specific hardware and input data.

Visual attention has a long history, and we recommend the following recent articles to the interested reader to learn more about the subject \cite{Li2018, Borji2019, Wang2021}. However, to really put it into practice, it is better to look for information about benchmarks \cite{Borji2012, Borji2015, Li2014}. In the present work, we select {\it Li et al.} since it provides an extensive evaluation of fixation prediction and salient object segmentation algorithms as well as statistics of major datasets. They provide a framework focusing on the performance of GBVS against several state-of-the-art proposals. The study also explains how to adapt fixation prediction algorithms to salient object detection by incorporating a segmentation stage. Fixation prediction algorithms target at predicting where people look in images. Salient object detection focus on a wide range of object-level computer vision applications. Since fixation prediction originated from cognitive and psychological communities, the goal is to understand the biological mechanism. Salient object detection does not necessarily need to understand biological phenomena.

Regarding the second level of explanation (algorithmic transparency) or the knowledge of the visual processing coding. We can observe two different approaches to incorporate learning into such study. The first is exemplified by a deep learning technique (DHSNET--Deep Hierarchical Saliency Network) since it is used as a building block in \cite{Contreras-Cruz2019}. This method is a fully convolutional network (FCN)-based method and it was designed to address the limitations of multi-layer perceptron (MLP)-based methods \cite{Liu2016}. FCN architectures leads to end-to-end spatial saliency representation learning and fast saliency prediction, within a single feed-forward process. FCN-based methods are now dominant in the field of computer vision. 

The second methodology is represented by evolutionary computation applying genetic programming. We identify two representative works. In \cite{Contreras-Cruz2019} the contribution is oriented towards the automatic design of combination models by using genetic programming. The proposed approach automatically selects the algorithms to be combined and the combination operators uses as input a set of candidate saliency detection methods and a set of combination operators. This idea follows a long history in computer vision about combination models. To achieve good results, authors rely on complex algorithms like DHSNET, using it as building blocks to the detriment of transparency since the method does not enhance the complex algorithms in the function set but only the output.

Since fixation prediction algorithms are complex heuristics another alternative is to work directly with some key parts of the algorithm to attempt to improve/discover the whole design. In \cite{Dozal2014} genetic programming serves to generate visual attention models--fixation prediction algorithms--to tackle salient object segmentation. However, the authors took a step back, returning to the first stage--theoretical transparency--and revisited {\it Koch et al.} looking for a suitable model susceptible to optimization \cite{Koch1985}. They develop an optimization-based approach to learn the complete model using a basic algorithm that serves the purpose of a template. This algorithm uses as a foundation the code reported in \cite{Itti1998}. In this way, {\it Dozal et al.} attempt to fulfill the second stage--algorithmic transparency--since they contemplate the difficulty of articulating the whole design exposed by Treisman and Gelade. To sum up, it is not easy to delegate all practical aspects to the computer according to the genetic programming paradigm. This way of looking for visual attention programs has already impacted practical applications like visual tracking \cite{Olague2018, Olague2019}. This method searches for new alternatives in the processes described in the feature integration theory (FIT). That includes processes for the acquisition of visual features, the computation of conspicuity maps, and the integration of features. Nevertheless, a drawback is that the visual attention models are evolved to detect a particular/single object in the image. 

In this work, we would like to identify all foreground regions and separate them from the background. Note that the foreground can contain any object on a particular database. This was the problem approached by {\it Contreras et al.} and is known as salient object detection. The idea is to replace Itti's algorithm with the proposal published in \cite{GBVS}  and the further adaptation and benchmark described in \cite{Li2014}.

Koch and coworkers adapt Treisman and Gelade's theory into basic computational reasoning. Itti's algorithm accomplishes two stages 1) visual feature acquisition and 2) feature integration. It consists of visual features extraction and computation of visual and conspicuity maps, feature combination, and the saliency map. GBVS is not different from Itti's implementation. However, it makes a better description of the technique through Markov processes. The idea is to adapt the GBVS algorithm to the symbolic framework of brain programming. Figure \ref{fig:evoGBVS} depicts the proposed algorithm where multiple functions are discovered through artificial evolution. GBVS is a graph-based bottom-up visual salience model. It consists of three steps: first, extraction of features from the image, second, creation of activation maps using the characteristic vectors, and third, normalization of activation maps and combining these maps into a master map. We adapt the algorithm described in \cite{Dozal2014} with the new proposal using four dimensions: color, orientation, shape, and intensity. In Koch's original work, there are three dimensions, each approached with a heuristic method, and the same for the integration step. We apply the set of functions and terminals provided in \cite{Clemente2015} with a few variants to discover optimal heuristic models for each of these stages. This algorithm uses the Markov chains to generate the activation maps. This approach is considered ``organic" because, biologically, individual ``nodes" (neurons) exist in a connected, retinotopically organized network (the visual cortex) and communicate with each other (synaptic activation) in a way that results in emergent behavior, including quick decisions about which areas of a scene require additional processing. 

\section{Methodology}


BP aims to emulate the behavior of the brain through an evolutionary paradigm using neuroscience knowledge for different vision problems. The first jobs to introduce this technique \cite{avc,Dozal2014} focused on automating the design of visual attention (VA) models and studied the way it surpasses previous human-made systems developed by VA experts. To perceive salient visual features, the natural dorsal stream in the brain has developed VA as a skill through selectivity and goal-driven behavior. The artificial dorsal stream (ADS) emulates this practice by automating acquisition and integration steps. Handy applications for this model are tracking objects from a video captured with a moving camera, as shown in \cite{Olague2018,Olague2019}.  


BP is a long process consisting of several stages summarized in two central ideas correlated with each other. First, the primary goal of BP is to discover functions that are capable of optimizing complex models by adjusting the operations within them. Second, a hierarchical structure inspired by the human visual cortex uses function composition to extract features from images. It is possible to adapt this model depending on the task at hand; \emph{e.g.}, the focus of attention can be applied to saliency problems \cite{Dozal2014}, or the complete artificial visual cortex (AVC) can be used for categorization/classification problems \cite{avc}. This study uses the ADS, explained to a full extent in the following subsections, to obtain as a final result the design of optimal salient object detection programs which satisfy the visual attention task.

\begin{table}
\centering
 \scalebox{0.70}{
    \begin{tabular}{|p{6cm}|p{5cm}|}
      \hline
      Functions for $EVO_O$   & Terminals for $EVO_O$ \\
      \hline
      $A+B$, $A-B$, $A \times B$, $A / B$, 
      $|A|$, $|A+B|$, $|A-B|$, $\log_2(A)$, $A / 2$, $A^2$, $\sqrt{A}$, $k\times A$, $A/k$, $A^{1/k}$, $A^{k}$, $(1/k) + A$, $A - (1/k)$, $G_{\sigma=1}(A)$, $G_{\sigma=2}(A)$, $D_{x}(A)$, 
      $D_{y}(A)$, $round(A)$, $\lfloor{A}\rfloor$, $\lceil A \rceil $, $inf(A, B)$, $sup(A, B)$, $thr(A)$, $AttenuateBorders(A)$, $ConvGabor(A)$ &
      $I_r$, $I_g$, $I_b$, $I_c$, $I_m$, $I_y$, $I_k$, $I_h$, $I_s$, $I_v$, 
      $D_{x}(I_{color})$, $D_{xx}(I_{color})$, $D_{y}(I_{color})$, $D_{yy}(I_{color})$, $D_{xy}(I_{color})$, $AttenuateBorders(I_{color})$, $ConvGabor(I_{color})$, \\
      \hline
      Functions for $EVO_C$   & Terminals for $EVO_C$ \\
      \hline
      $A+B$, $A-B$, $A \times B$, $A / B$, $\log_2(A)$, $exp(A)$, $|A|$, 
      $A^2$, $\sqrt{A}$, $(A)^c$, $thr(A)$, $round(A)$, $\lfloor{A}\rfloor$, $\lceil A \rceil $, $k\times A$, $A/k$, $A^{1/k}$, $A^{k}$, $(1/k) + A$, $A - (1/k)$ &  
      $I_r$, $I_g$, $I_b$, $I_c$, $I_m$, $I_y$, $I_k$, $I_h$, $I_s$, $I_v$,$DKL_{r}$, $DKL_{\Phi}$, $DKL_{\Theta}$, $Op_{r-g}(I)$,$Op_{b-y}(I)$ \\ 
      \hline
     Functions for $EVO_S$   & Terminals for $EVO_S$ \\
      \hline
      $A+B$, $A-B$, $A \times B$, $A / B$, $|A|$, $k\times A$, $A/k$, $A^{1/k}$, $A^{k}$, $(1/k) + A$, $A - (1/k)$,
      $round(A)$, $\lfloor A \rfloor $, $\lceil A \rceil $, $A \oplus SE_d$, 
      $A \oplus SE_s$, $A \oplus SE_{dm}$, $A\ominus SE_{d}$, $A\ominus SE_{s}$, 
      $A\ominus SE_{dm}$, $Sk(A)$, $Perim(A)$, $A\circledast SE_{d}$, $A\circledast SE_{s}$, 
      $A\circledast SE_{dm}$, $T_{hat}(A)$, $B_{hat}(A)$, $A\circledcirc SE_s$, $A\odot SE_s$, $thr(A)$    &
      $I_r$, $I_g$, $I_b$, $I_c$, $I_m$, $I_y$, $I_k$, $I_h$, $I_s$, $I_v$ \\ 
      \hline   
      Functions for $EFI$   & Terminals for $EFI$ \\
      \hline
      $A+B$, $A-B$, $A \times B$, $A / B$, $|A|$, $|A+B|$, $|A-B|$, $k\times A$, $A/k$, $A^{1/k}$, $A^{k}$, $(1/k) + A$, $A - (1/k)$,,$Hist(A)$, $round(A)$, $\lfloor A \rfloor $, $\lceil A \rceil $, $thr(A)$, $(A)^2$,
      $\sqrt{A}$, $exp(A)$, $G_{\sigma=1}(A)$, $G_{\sigma=2}(A)$, $D_{x}(A)$, $D_{y}(A)$  &
      $CM_d$, $D_{x}(CM_d)$, $D_{xx}(CM_d)$, $D_{y}(CM_d)$, $D_{yy}(CM_d)$, $D_{xy}(CM_d)$ \\
      \hline

     \end{tabular}
}
\caption{Functions and terminals for the ADS.}
\label{tab:Funciones}
\end{table}

\subsection{Initialization}
\label{sec:initialization}

BP begins with a randomized generation, along an evolutionary process defined by a set of initialization variables such as population size, size of solutions or individuals, or crossover-mutation probabilities. An individual represents a computer program written with a group of syntactic trees embedded into hierarchical structures. In this work, individuals within the population contain functions corresponding to one of the four available visual operators ($VO$). Table \ref{tab:Funciones} shows the list of functions and terminals used for each $VO$ or visual map ($VM$). The table includes arithmetic functions between two images $A$ and $B$, transcendental and square functions,  square root function, image complement, color opponencies (Red-Green and Blue-Yellow), dynamic threshold function, arithmetic functions between an image $A$ and a constant $k$. It also includes transcendental operations with a constant $k$ and the spherical coordinates of the DKL color space. The table also incorporates round, half, floor, and ceil functions over an image $A$, dilation and erosion operators with the disk, square, and diamond structure element ($SE$), skeleton operator over the image $A$, find the perimeter of objects in the image $A$, hit or miss transformation with the disk, square, and diamond structures. Also, we include morphological top-hat and bottom-hat filtering over the image $A$, opening and closing morphological operator on $A$, absolute value applied to $A$, and the addition and subtraction operators. Finally, we add the infimum and supremum functions between images $A$ and $B$, the convolution of the image $A$, a Gaussian filter with $\sigma = 1$ or $2$, and derivative of the image $A$ along direction $x$ and $y$. Note that all dimensions include elementary functions since these can be employed to compose high-level properties (invariance to rotation, translation, scaling, and illumination ) through the usage of the template design.

\subsection{Individual Representation}

\label{sec:initialization}


We represent individuals by using a set of functions for each $VO$ defined in Section \ref{ADS}. Entities are encoded into a multi-tree architecture and optimized through evolutionary operations of crossover and mutation.


The architecture uses four syntactic trees, one for each evolutionary visual operator ($EVO_O$, $EVO_C$, $EVO_S$) regarding orientation, color, and shape. We then merge the $CMs$ produced by the center-surround process--including feature and activation maps--using an EFI tree, generating a saliency map (SM) as a result. Section \ref{sec:features} provides details about the usage of these $EVOs$; additionally, Figure \ref{fig:evoGBVS} provides a graphical representation of the complete BP workflow. After initializing the first generation of individuals, the fitness of each solution is tested and used for creating a new population.

\subsection{Artificial Dorsal Stream}
\label{ADS}




The ADS models some components of the human visual cortex, where each layer represents a function achieved by synthesis through a set of mathematical operations; this constitutes a virtual bundle. We select visual features from the image to build an abstract representation of the object of interest. Therefore, the system looks for salient points (at different dimensions) in the image to construct a saliency map used in the detection process. The ADS comprises two main stages: the first acquires and transforms features in parallel that highlight the object, while in the second stage, all integrated features serve the goal of object detection.

\subsubsection{Acquisition and Transformation of Features}
\label{sec:features}

In this stage, different parts of the artificial brain automatically separate basic features into dimensions. The entrance to the ADS is a color image $ I $ defined as the graph of a function as follows.

\textbf{Definition 1. Image as the graph of a function}. \textit{Let $f$ be a function $f:U \subset \mathbb{R}^2 \rightarrow \mathbb{R}$. The graph or image $I$ of $f$ is the subset of $\mathbb{R}^3$ that consist of the points $(x, y, f(x,y))$, in which the ordered pair $(x,y)$ is a point in $U$ and $f(x,y)$ is the value at that point. Symbolically, the image $I = \{(x,y,f(x,y)) \in \mathbb{R}^3 | (x,y) \in U\}$}.


From this definition, we can highlight how images are variations in light intensity along the two-dimensional plane of camera sensors. Regarding visual processing for feature extraction of the input image, we consider multiple color channels to build the set $I_{color}$ = $\{I_r$, $I_g$, $I_b$, $I_c$, $I_m$, $I_y$, $I_k$, $I_h$, $I_s$, $I_v\}$, where each element corresponds to the color components of the RGB (red, green, blue), HSV (Hue, Saturation and Value) and CMYK (Cyan, Magenta, Yellow and black) color spaces. We define the optimization process through the formulation of an appropriate search space and evaluation functions.

\subsubsection{Feature Dimensions}


In this step, we obtain relevant characteristics from the image by decomposing it and analyzing key features. Three $EVOs$ transform the input picture $I_{color }$ through each $VO$ defined as $EVO_d: I_{color}\rightarrow VM_d$ and applied in parallel to emphasize specific characteristics of the object. Note that the fourth $VM_{Int}$ is not evolved and is calculated with the average of the RGB color bands. These $EVOs$ are operators generated in Section \ref{sec:initialization}. Individuals-programs represent possible configurations for feature extraction that describe input images and are optimized using the evolutionary process. We perform these transformations to recreate the process of extracting information following the FIT. When applying each operator, a $VM$ generated for each dimension represents a partial procedure within the overall process. Each $VM$ is a topographic map that represents, in some way, an elementary characteristic of the image.

\subsection{Creating the Activation Maps}
After selecting the visual operators generated by the evolutionary process the feature maps complete the feature extraction. Next, activation maps are created as follows. Suppose we are given a feature map \(M:[n]^{2} \rightarrow \mathbb{R}\) the goal is to compute an activation map for each dimension \(A:[n]^{2} \rightarrow \mathbb{R}\) such that locations \((i,j) \in [n]^{2}\) on the image, or as a proxy, \(M(i,j)\), is somehow unusual in its neighborhood will correspond to high values of activation $A$. The dissimilarity of \(M(i,j)\) and \(M(p,q)\) is given by:
\begin{equation}
    d((i,j)||(p,q)) \triangleq \log \displaystyle \frac{M(i,j)}{M(p,q)} \; .
\end{equation}

\subsection{A Markovian Approach}
Now, consider a fully connected graph denoted as \(G_{A}\). For each node $M$ with its indexes \((i,j) \in [n]^{2}\) connected to the other nodes. The edge point of a node in the two-dimensional plane \((i,j)\) to the node \((p,q)\) will be the weight and is defined as follows:

\begin{equation}
    w_{1}((i,j),(p,q)) \triangleq d((i,j)||(p,q)) \cdot F(i-p,j-q) \; ,
\end{equation}
where
\begin{equation}
    F(a,b) \triangleq exp \displaystyle \frac{-(a^{2}+b^{2})}{2\sigma^{2}} \; ,
\end{equation}
and \(\sigma\) is a free parameter of the algorithm. Thus, the weight of the edge from node \((i,j)\) to node \((p,q)\) is proportional to their dissimilarity and their closeness in the domain of M. It is possible then to define a Markov chain on \(G_{A}\) by normalizing the weights of the outbound edges of each node to 1, and drawing an equivalence between nodes-states, and edges weights-transition probabilities.

\subsection{Normalizing an Activation Map}
This step is crucial to any saliency algorithm and remains a rich area of study. GBVS proposes another Markovian algorithm and the goal of this step is a mass-concentration in the activation maps. Authors construct a graph \(G_{N}\) with \(n^{2}\) nodes labeled with indices from \([n]^{2}\). For each node \((i,j)\) and \((p,q)\) connected, they introduce an edge from

\begin{equation}
    w_{2}((i,j),(p,q)) \triangleq A(p,q) \cdot F(i-p,j-q) \; .
\end{equation}
\noindent
Once again, each node's output edges are normalized to unity, and treating the resulting graph as a Markov chain makes it possible to calculate the equilibrium distribution over the nodes. The mass will flow preferentially to those nodes with high activation. The artificial evolutionary process works with the modified version of GBVS. To improve the results, we can add the MCG during the evolution or after since the image segmentation computational cost with this algorithm is very high.

\subsection{Genetic Operations}

We follow the approach detailed in \cite{Dozal2014} where the template represents an individual containing a set of $VMs$ coded into an array of trees similar to a chromosome and where each visual operator within the chromosome is a gene code. In other words, the chromosome is a list of visual operators ($VOs$), and each $VO$ is a gene. Therefore, we apply four genetic operators:

\begin{itemize}
\item Chromosome-level crossover. The algorithm randomly selects a crossing point from the list of trees. The process built a new offspring by the union of the left section of the first parent with the right section of the second parent. 

\item Gene level crossover: This operator selects two $VOs$, chosen randomly, from a list of trees, and for each function of both trees (genes), the system chooses a crossing point randomly, then the sub-trees under the crossing point are exchanged to generate two new visual operators. Therefore, this operation creates two new children-chromosomes.

\item Chromosome-level mutation: The algorithm randomly selects a mutation point within a parent's chromosome and replaces the chosen operator completely with a randomly generated operator.

\item Gene-level mutation: Within a visual operator, randomly chosen, the algorithm selects a node, and the mutation operation randomly alters the sub-tree that results below this point.
\end{itemize}

Once we generate the new population, the evolutionary process continues, and we proceed to evaluate the new offspring. 

\subsection{Evaluation Measures}
\label{sec:scores}

Evolutionary algorithms usually apply a previously defined fitness function to evaluate the individuals' performance. BP designs algorithms using the generated $EVOs$ to extract features from input images through the ADS hierarchical structure depicted in Figure \ref{fig:evoGBVS}. Experts agree on the way to evaluate the various proposals for solutions to the problem of salient object detection. In this work, we follow the protocol detailed with source code in \cite{Borji2015} and apply two main evaluation measures: Precision-Recall and F-measure. The first is given through Equation (\ref{PR}):

\begin{equation} 
Precision = \frac {|BM\cap G|}{|BM|}, \quad Recall = \frac {|BM\cap  G|}{|G|} \; .
\label{PR}
\end{equation}

\noindent
To compute a saliency map $S$, we convert it to a binary mask $BM$, and compute $Precision$ and $Recall$ by comparing $BM$ with ground-truth $G$. In this definition binarization is a key step in the evaluation. The benchmark offers a method based on thresholds to generate a precision-recall curve. The second measure (Equation (\ref{F-measure})) is made with this information to obtain a figure of merit:

\begin{equation} 
F_{\beta } = \frac {(1+\beta ^{2}) Precision \times Recall}{\beta ^{2} Precision + Recall} \; .
\label{F-measure}
\end{equation}
\noindent
This expression comprehensively evaluate the quality of a saliency map. The F-measure is the weighted harmonic mean of precision and recall. In the benchmark $\beta^2$ is set to $0.3$ to increase the importance of the precision value.

We calculate both evaluations with two variants. In our first approach, we obtain the maximum F-measure considering different thresholds for each image during the binarization process. Then, we calculate the average of all photos in the training or testing set; see \cite{Dozal2014}. The second variant is the one used in the benchmark, which consists of first calculating the average that results from varying the thresholds and then reporting the maximum resulting from evaluating all the images. We will use both approaches during the experiments.

\begin{table}[H]

    \centering

    \begin{tabular}{c c}

         \toprule

         \textbf{Parameters} & \textbf{Description} \\

         \midrule

         Generations & 30 \\

         Population size & 30 \\

         Initialization & Ramped half-and-half \\

         Crossover at chromosome level & 0.8 \\

         Crossover at gene level & 0.8 \\

         Mutation at chromosome level & 0.2 \\

         Mutation at gene level & 0.2 \\

         Tree depth & Dynamic depth selection \\

         Dynamic maximum depth & 7 levels \\

         Maximum actual depth & 9 levels \\

         Selection & Tournament selection \\ & with lexicographic \\ & parsimony pressure. \\

         Elitism & Keep the best individual \\

         \toprule

    \end{tabular}

    \caption{Main parameters settings of the BP algorithm.}

    \label{tab:comp-results}

\end{table}

\section{Experiments and Results}

Designing machine learning systems requires the definition of three different components: algorithm, data, and measure. In this section, we evaluate the proposed evolutionary algorithm with a standard test. Thus, the goal is to benchmark our algorithm against external criteria. In this way,  we need to run a series of tests based on data and measures provided by well-known experts. Finally, we contrast our results with several algorithms in the state-of-the-art. 

In this research, we follow the protocol detailed in \cite{Li2014}. This benchmark is of great help because it gives us the possibility of accessing the source code of various algorithms to make a more exhaustive comparison. This benchmark also analyzes blunt flaws in the design of salience benchmarks known as database design bias produced by emphasizing stereotypical salience concepts. This benchmark makes an extensive evaluation of fixation prediction and salient object segmentation algorithms. We focus on the salient object detection part, consisting of three databases FT, IMGSAL, and PASCAL-S. We present complete results next. Also, we include a test of the best program with the databases proposed in \cite{Contreras-Cruz2019}.



\subsection{Image Databases}
FT is a database with 800 images for training and 200 images for testing. Authors of the benchmark reserve this last set of 200 images for comparison. We use the training dataset to perform a k-fold technique ($k=5$) to find the best individual. The training dataset is randomly partitioned into five subsets of 160 images. A single subset of the $k$ subsets is retained as the validation data for testing the model, and the remaining $k - 1$ subsets are used as training data. Table \ref{tab:ft-results-gbvs-mcg} reports the best program results for 30 executions in the k-fold technique. Each execution was run considering the parameters from Table \ref{tab:comp-results}. We follow the same procedure in the PASCAL-S database, and the final results are given in Figure \cite{Everingham2010} while adding ocular fixation information and object segmentation labeling. The training set consists of 680 images, while the test set consists of 170 images for comparison. We divide the training set into 5 subsets of 136 images to perform 5-fold cross-validation to discover the best algorithm. Finally, we run the experiments on the IMGSAL dataset, which contains 235 images. The database splits into 188 images for training and 47 images for testing. We select 185 out of the training set and split it into five subsets of 37 images to perform 5-fold cross-validation.

FT contains a diverse representation of animate and inanimate objects with image sizes ranging from $324 \times 216$ up to $400 \times 300$. PASCAL-S contains scenes of domestic animals, persons, means of air and sea transport with image sizes ranging from $200 \times 300$ up to $375 \times 500$. Finally, IMGSAL has wild animals and flora and different objects and persons with image sizes of $480 \times 640$. All images in the three datasets came with their corresponding ground truth. The manual segmentation was carefully made in FT and PASCAL-S datasets to obtain accurate ground truth, while IMGSAL provides a ground truth that was purposefully segmented following a raw segmentation to illustrate a real case where humans indicate an object's location inaccurately.

\subsection{Experiments with Dozal's Fitness Function on GBVSBP}

\begin{table}[H]
    \centering
    \resizebox{\textwidth}{!}{
        \begin{tabular}{*{13}{c}}
        \toprule
        \multicolumn{13}{c}{\textbf{FT}} \\
        Fold & \multicolumn{2}{c}{Run 1} & \multicolumn{2}{c}{Run 2} & \multicolumn{2}{c}{Run 3} & \multicolumn{2}{c}{Run 4} & \multicolumn{2}{c}{Run 5} & \multicolumn{2}{c}{Run 6} \\ 
        \midrule
         & Trng & Test & Trng & Test & Trng & Test & Trng & Test & Trng & Test & Trng & Test\\ 
            1 & 75.34 & 76.10 & 75.07 & 73.97 & 77.57 & 76.75 & 76.98 & 75.29 & 77.34 & 75.74 & 77.08 & 75.36 \\ 
            2 & 74.88 & 71.60 & 72.92 & 69.81 & 75.10 & 72.97 & 74.65 & 71.27 & 73.62 & 71.87 & 76.41 & 72.25 \\ 
            3 & \textbf{78.18} & 76.58 & 76.69 & 71.77 & 77.03 & 72.97 & 76.75 & 72.18 & 76.68 & 72.39 & 77.01 & 72.21 \\ 
            4 & 75.30 & 77.05 & 74.78 & 74.25 & 74.90 & 71.47 & 75.86 & \textbf{77.67} & 74.60 & 76.13 & 77.03 & 74.98 \\ 
            5 & 75.15 & 74.54 & 73.14 & 73.45 & 74.74 & 74.97 & 74.62 & 74.17 & 75.18 & 75.06 & 76.09 & 74.37 \\ 
            Average & 75,77 & \textbf{75.17} & 74,52 & 72.32 & 75,87 & 73.83 & 75,77 & 74.12 & 75,48 & 74.24 & \textbf{76,72} & 73.83 \\
            $\sigma$ & 1.36 & 2.21 & 1.54 & 1.97 & 1.33 & 2.05 & 1.12 & 2.54 & 1.52 & 1.97 & \textbf{0.45} & \textbf{1.51}\\ 
        \midrule
    \end{tabular}
    }
    \caption{Performance of the best individuals of the GBVSBP model for all \textbf{FT} database runs.}
    \label{tab:ft-results-gbvs}
\end{table}

Table \ref{tab:ft-results-gbvs} details the experimental results after applying the k-fold cross-validation method on the FT database. As we can observe, BP optimizes the model scoring the highest value of $78.18\%$ for the best program during training at the forth run and the forth fold, while achieving $77.67\%$ during testing also in the first run but at the fourth fold. $\sigma$ remains low during the experimental runs. BP scores its highest value $\sigma = 2.54$ in the fourth testing run while achieving $\sigma = 0.45$ and $\sigma = 1.51$ during the sixth run for training and testing, respectively. On average, the methodology scores its highest fitness of $76.72\%$ for training and $75.17\%$ for testing. Table \ref{tab:ft-best-ind-estructure} presents the best solution. The selected program corresponds to the training stage for this kind of table.

\begin{table}[H]
    \centering
    \resizebox{\textwidth}{!}{
        \begin{tabular}{*{13}{c}}
            \hline
           \textbf{$EVO_{d}$ and $EFI$ operators} & \textbf{Fitness} \\
            \hline
            \multirow{4}{20cm}{$EVO_{O} = I_{k}$ \\ $EVO_{C} = I_{b} + 1.00$ \\ $EVO_{S} = top - hat(I_{m}$ x $0.31)$ \\ $EFI = (CM_{C})^{2^{2}}$} & Training = 0.7818 \\ & Testing = 0.7767 \\ & \\ & \\
            \hline
        \end{tabular}
    }
    \caption{Structure of the operators corresponding to the best solution for the \textbf{FT} database.}
    \label{tab:ft-best-ind-estructure}
\end{table}

\begin{table}[H]
    \centering
    \resizebox{\textwidth}{!}{
        \begin{tabular}{*{13}{c}}
        \toprule
        \multicolumn{13}{c}{\textbf{IMGSAL}} \\
        Fold & \multicolumn{2}{c}{Run 1} & \multicolumn{2}{c}{Run 2} & \multicolumn{2}{c}{Run 3} & \multicolumn{2}{c}{Run 4} & \multicolumn{2}{c}{Run 5} & \multicolumn{2}{c}{Run 6} \\ 
        \midrule
         & Trng & Test & Trng & Test & Trng & Test & Trng & Test & Trng & Test & Trng & Test\\ 
            1 & 68.30 & 66.88 & 67.62 & 65.29 &  {68.86} &  {64.28} &  {\textbf{72.49}} &  {68.66} &  {66.23} &  {65.45} &  {59.24} &  {62.2} \\ 
            2 & 65.23 & 64.06 & 67.10 & 64.76 &  {66.79} &  {64.82} &  {64.18} &  {70.12} &  {69.25} &  {69.69} &  {66.26} &  {64.96} \\ 
            3 & 68.86 & 66.25 & 64.99 & 62.73 &  {60.84} &  {61.82} &
             {64.39} &  {65.26} &  {68.22} &  {63.93} &  {66.7} &   {62.42} \\ 
            4 & 62.72 & \textbf{70.91} & 57.37 & 61.27 &  {68.42} &  {67.16} &  {66.52} &  {66.87} &  {68.52} &  {67.17} &  {61.87} &  {64.56} \\ 
            5 & 70.74 & 67.10 & 68.99 & 65.43 & 66.26 & 60.37 &  {62.86} &  {62.73} &  {70.36} &  {66.6} &  {58.94} &  {61.45} \\ 
            Average & 67.17 & \textbf{67.04} & 65.21 & 63.90 & 66.23 & 63.69 & 66.09 & 66.73 & \textbf{68.52} & 66.57 & 62.6 & 63.12 \\
            $\sigma$ & 3.18 & 2.48 & 4.61 & 1.82 & 3.20 & 2.65 & 3.81 & 2.89 & \textbf{1.52} & 2.14 & 3.72 & \textbf{1.55}\\ 
        \midrule
    \end{tabular}
    }
    \caption{Performance of the best individuals of the GBVSBP model for all \textbf{IMGSAL} database runs.}
    \label{tab:imgsal-results-gbvs}
\end{table}

Table \ref{tab:imgsal-results-gbvs} details the experimental results after applying the k-fold cross-validation method on the IMGSAL database. The fitness function results on this dataset show high variability since, in training, the results range from $57.37\%$ in the second run--up to $70.74\%$ in the first run. The high dispersion is due to the complexity of the problem generated by the poor manual segmentation. The experiment shows an average oscillating between $62.60\%$ and $68.52\%$ for training with $\sigma = 3.72$ and $\sigma = 1.52$ respectively. On the other hand, during testing, the algorithm highest score in average is $67.04\%$ with $\sigma = 2.48$. Table \ref{tab:imgsal-best-ind-estructure} reports the best solution.

\begin{table}[H]
    \centering
    \resizebox{\textwidth}{!}{
        \begin{tabular}{*{13}{c}}
            \hline
           \textbf{$EVO_{d}$ and $EFI$ operators} & \textbf{Fitness} \\
            \hline
            \multirow{4}{20cm}{$EVO_{O} = ((G_{(\sigma = 1)}(G_{(\sigma = 1)}(I_{v})) - 0.44) - 0.69)$ \\ $EVO_{C} = (I_{c}^{0.44})$ \\ $EVO_{S} = I_{m}$ \\ $EFI = G_{(\sigma = 2)}(D_{y}(D_{y}(CM_{C})))$} & Training = 0.7249 \\ & Testing = 0.7091 \\ & \\ & \\
            \hline
        \end{tabular}
    }
    \caption{Structure of the operators corresponding to the best solution for the \textbf{IMGSAL} database.}
    \label{tab:imgsal-best-ind-estructure}
\end{table}

\begin{table}[H]
    \centering
    \resizebox{\textwidth}{!}{
        \begin{tabular}{*{13}{c}}
        \toprule
        \multicolumn{13}{c}{\textbf{PASCAL-S}} \\
        Fold & \multicolumn{2}{c}{Run 1} & \multicolumn{2}{c}{Run 2} & \multicolumn{2}{c}{Run 3} & \multicolumn{2}{c}{Run 4} & \multicolumn{2}{c}{Run 5} & \multicolumn{2}{c}{Run 6} \\ 
        \toprule
         & Trng & Test & Trng & Test & Trng & Test & Trng & Test & Trng & Test & Trng & Test\\ 
            1 & 63.58 & 62.31 & 62.92 & 61.83 & 65.19 & 65.33 & 62.27 & 61.34 & 62.84 & 63.12 & 63.79 & 62.25 \\ 
            2 & 61.53 & 58.92 & 64.70 & 57.51 & 63.02 & 56.88 & 64.64 & 59.80 & 65.29 & 58.81 & 64.50 & 58.57 \\ 
            3 & 61.58 & 64.61 & 63.47 & 65.46 & 62.18 & 64.13 & 63.71 & 65.22 & 63.20 & \textbf{67.66} & 59.71 & 64.90 \\ 
            4 & 59.60 & 62.82 & 58.20 & 59.77 & 59.23 & 60.87 & 60.45 & 63.68 & 59.72 & 61.78 & 59.84 & 60.63 \\ 
            5 & 63.20 & 65.81 & 64.46 & 63.25 & 63.90 & 63.47 & 66.03 & 64.55 & 66.22 & 66.79 & \textbf{66.39} & 66.78 \\ 
            Average & 61.90 & 62.89 & 62.75 & 61.56 & 62.70 & 62.14 & 63.42 & 62.92 & \textbf{63.45} & \textbf{63.63} & 62.85 & 62.83 \\
            $\sigma$ & \textbf{1.58} & 2.63 & 2.64 & 3.07 & 2.24 & 3.36 & 2.15 & \textbf{2.28} & 2.52 & 3.64 & 2.96 & 2.99 \\ 
        \toprule
    \end{tabular}
    }
    \caption{Performance of the best individuals of the GBVSBP model for all \textbf{PASCAL-S} database runs.}
    \label{tab:pascal-results-gbvs}
\end{table}

The experimental results of the GBVSBP Model with the PASCAL-S database show excellent stability as seen in Table \ref{tab:pascal-results-gbvs} with a low standard deviation, especially in training with the first run-scoring 1.58.  During training, fitness reaches highest in the fifth fold of run 6, scoring $66.39\%$, while on average, the fifth run scores first place with $63.45\%$. Regarding the testing stage, the algorithm scores the best individual at the fifth fold with a $67.66\%$, and on average, the best run was the fifth with $63.63\%$. Table \ref{tab:pascal-best-ind-estructure} presents the best set of trees.

\begin{table}[H]
    \centering
    \resizebox{\textwidth}{!}{
        \begin{tabular}{*{13}{c}}
            \hline
           \textbf{$EVO_{d}$ and $EFI$ operators} & \textbf{Fitness} \\
            \hline
            \multirow{4}{20cm}{$EVO_{O} = \lfloor(G_{(\sigma = 1)}(D_{y}(I_{s})))\rfloor$ \\ $EVO_{C} = (((((I_{m} - 0.62) - 0.62) + I_{y}) + (I_{y} + I_{v})) + (I_{y} + I_{v}))$ \\ $EVO_{S} = I_{m}$ \\ $EFI = D_{y}(D_{y}(CM_{C}))$} & Training = 0.6639 \\ & Testing = 0.6766 \\ & \\ & \\
            \hline
        \end{tabular}
    }
    \caption{Structure of the operators corresponding to the best solution for the \textbf{PASCAL-S} database.}
    \label{tab:pascal-best-ind-estructure}
\end{table}

\begin{table*}
    \centering
    \resizebox{\textwidth}{!} 
{
        \begin{tabular}{*{13}{c}}
           \toprule
            \multicolumn{13}{c}{\textbf{FT database}} \\
            Fold & \multicolumn{2}{c}{Run 1} & \multicolumn{2}{c}{Run 2} & \multicolumn{2}{c}{Run 3} & \multicolumn{2}{c}{Run 4} & \multicolumn{2}{c}{Run 5} & \multicolumn{2}{c}{Run 6} \\ 
           \midrule
             & Trng & Test & Trng & Test & Trng & Test & Trng & Test & Trng & Test & Trng & Test\\ 
            
                1 & 93.39 & 92.90 & 93.62 & 93.09 & 93.67 & 93.35 & 93.29 & 94.14 & 93.55 & 92.19 & 93.27 & 93.44\\ 
            
                2 & \textbf{94.53} & 86.67 & 94.02 & 86.12 & 93.75 & 84.61 & 93.73 & 86.67 & 93.80 & 86.84 & 93.74 & 85.38\\ 
            
                3 & 92.84 & 93.55 & 92.82 & 92.76 & 92.69 & 92.22 & 93.05 & 94.36 & 92.45 & 92.78 & 92.63 & 92.90 \\ 
            
                4 & 93.33 & 92.77 & 92.86  & 93.64 & 92.76 & 93.54 & 92.98 & 91.59 & 92.87 & 93.89 & 93.23 & 90.62\\ 
            
                5 & 93.25 & 92.86 & 93.45 & \textbf{95.06} & 93.02 & 94.31 & 93.20 & 92.22 & 93.03 & 93.31 & 93.00 & 92.18\\ 
            
                Average & \textbf{93.47} & 91.75 & 93.35 & \textbf{92.13} & 93.18 & 91.61 & 93.25 & 91.80 & 93.14 & 91.80 & 93.17 & 90.90 \\ 
            
                $\sigma$ & 0.63 & 2.86 & 0.51 & 3.48 & 0.50 & 3.98 & \textbf{0.29} & 3.11 & 0.54 & \textbf{2.84 }& 0.41 & 3.26\\ 
            \toprule
        \end{tabular}
}
    \caption{Performance of the best individuals of the GBVSBP + MCG model for all \textbf{FT} database runs.}
    \label{tab:ft-results-gbvs-mcg}
\end{table*}

The experiment with our second model GBVSBP+MCG and the FT database shows outstanding results compared to the previous model, see Table \ref{tab:ft-results-gbvs-mcg}. Another remarkable difference is the stability during training regarding the standard deviation, whose performance descend a little while testing the best models, where four values score above $\sigma = 3$. Meanwhile, in the testing stage, the best individual achieves $95.06\%$ in the second run. On average, the algorithm discovered the best individuals considering all folds in the first run with $93.47\%$, while the second run reports the best results with an average of $92.13\%$. Table \ref{tab:ft-mcg-best-ind-estructure} shows the best solution.

\begin{table}[H]
    \centering
    \resizebox{\textwidth}{!}{
        \begin{tabular}{*{13}{c}}
            \hline
           \textbf{$EVO_{d}$ and $EFI$ operators} & \textbf{Fitness} \\
            \hline
            \multirow{4}{20cm}{$EVO_{O} = G_{\sigma = 1}(I_{m})$\\ $EVO_{C} = Complement(Complement(DKL_{r}))$\\$EVO_{S} = ((bottom - hat(I_{b})$ x $(bottom - hat(I_{b}))$\\$EFI = |(G_{(\sigma = 1)}(CM_{MM})$ x $0.63) + ((D_{y}(D_{y}(CM_{C})) - D_{y}(D_{y}(CM_{MM}))) - D_{y}(D_{y}(CM_{MM})))|$} & Training = 0.9453 \\ & Testing = 0.9506 \\ & \\ & \\ & \\ & \\ & \\
            \hline
        \end{tabular}
    }
    \caption{Structure of the operators corresponding to the best solution for the \textbf{FT} database.}
    \label{tab:ft-mcg-best-ind-estructure}
\end{table}

\subsection{Experiments with Benchmark's Score}

As the second round of experiments, we adapt the algorithm to use as a fitness function the proposed benchmark's score as explained earlier, see Section \ref{sec:scores}. From now on, all report experiments considered this way of evaluation. Table \ref{tab:ft-results-gbvs-new}
provides the results of the k-fold experimentation considering the GBVSBP algorithm with the FT database, while Table \ref{tab:ft-results-gbvs-mcg-new} provides partial results with the GBVSBP+MCG model to illustrate the performance. As can be seen, there is a decrease, but the ranking remains unaltered, as we verified with the solutions. The results considering the datasets IMGSAL and PASCAL-S are in Tables \ref{tab:imgsal-results-gbvs-new} and \ref{tab:pascal-results-gbvs-new}, respectively.

\begin{table}[H]
    \centering
    \resizebox{\textwidth}{!}{
        \begin{tabular}{*{13}{c}}
            \toprule
            \multicolumn{13}{c}{\textbf{FT}} \\
            Fold & \multicolumn{2}{c}{Run 1} & \multicolumn{2}{c}{Run 2} & \multicolumn{2}{c}{Run 3} & \multicolumn{2}{c}{Run 4} & \multicolumn{2}{c}{Run 5} & \multicolumn{2}{c}{Run 6} \\ 
            \midrule
             & Trng & Test & Trng & Test & Trng & Test & Trng & Test & Trng & Test & Trng & Test\\ 
                1 & 71.27 & 71.94 & 70.23 & 70.39 & 72.04 & 72.23 & 70.73 & 70.99 & 71.22 & 70.80 & \textbf{72.59} & 73.65 \\
                2 & 70.00 & 63.05 & 67.04 & 65.73 & 68.66 & 68.86 & 67.57 & 69.01 & 69.5 & 66.58 & 68.18 & 67.78 \\ 
                3 & 71.71 & 64.82 & 71.81 & 67.92 & 71.16 & 68.15 & 71.22 & 64.15 & 72.14 & 68.02 & 71.08 & 66.86\\ 
                4 & 69.23 & 73.45 & 70.69 & 66.13 & 71.32 & 68.11 & 68.15 & \textbf{73.81} & 68.77 & 71.13 & 70.87 & 69.42\\ 
                5 & 68.08 & 67.68 & 68.52 & 67.20 & 68.88 & 66.03 & 69.38 & 64.25 & 69.17 & 68.19 & 68.73 & 67.59\\ 
                Average & 70.06 & 68,19 & 69.66 & 67,47 & 70.41 & 68,68 & 69.41 & 68,44 & 70.07 & 68,94 & \textbf{70.29} & \textbf{69,06}\\
                $\sigma$ & \textbf{1.48} & 4,47 & 1.88 & \textbf{1,85} & 1.54 & 2,25 & 1.58 & 4,23 & 1.51 & 1,95 & 1.81 & 2,73\\ 
            \midrule
        \end{tabular}
    }
    \caption{Performance of the best individuals of the GBVSBP model for all \textbf{FT} database runs.}
    \label{tab:ft-results-gbvs-new}
\end{table}

With this new F-measure, the results obtained show greater stability globally according to Table \ref{tab:ft-results-gbvs-new} despite runs 1 and 4 reporting $\sigma = 4.47$ and $\sigma = 4,23$. All other values are below $2\%$. The fitness of the individuals, despite that a decrease is observed, remains with competitive results as the best individual in the training stage achieves $72.59\%$, and the best in the test stage reaches $73.81\%$. In this experiment, the highest average fitness was reflected in the last run considering all folds and for both stages with values of $70.29\%$ and $69.06\%$. Table \ref{tab:ft-new-best-ind-estructure} gives the best set of visual operators.

\begin{table}[H]
    \centering
    \resizebox{\textwidth}{!}{
        \begin{tabular}{*{13}{c}}
            \hline
           \textbf{$EVO_{d}$ and $EFI$ operators} & \textbf{Fitness} \\
            \hline
            \multirow{4}{20cm}{$EVO_{O} = I_{r}$\\ $EVO_{C} = (((I_{k} + I_{m})+I_{m})+I_{m})$\\$EVO_{S} = threshold(I_{y})$\\$EFI = D_{x}(D_{y}(CM_{S}))$} & Training = 0.7259 \\ & Testing = 0.7381 \\ & \\ & \\
            \hline
        \end{tabular}
    }
    \caption{Structure of the operators corresponding to the best solution for the \textbf{FT} database.}
    \label{tab:ft-new-best-ind-estructure}
\end{table}

\begin{table}[H]
    \centering
    \resizebox{\textwidth}{!}{
        \begin{tabular}{*{13}{c}}
        \toprule
        \multicolumn{13}{c}{\textbf{IMGSAL}} \\
        Fold & \multicolumn{2}{c}{Run 1} & \multicolumn{2}{c}{Run 2} & \multicolumn{2}{c}{Run 3} & \multicolumn{2}{c}{Run 4} & \multicolumn{2}{c}{Run 5} & \multicolumn{2}{c}{Run 6} \\ 
        \midrule
         & Trng & Test & Trng & Test & Trng & Test & Trng & Test & Trng & Test & Trng & Test\\ 
            1 & 57.19 & 60.16 & 60.09 & 60.17 & \textbf{65.35} & \textbf{63.45} &  {62.05} &  {60.31} &  {65.09} &  {62.20} &  {60.24} &  {60.8} \\ 
            2 & 56.12 & 59.94 & 55.55 & 60.25 & 63.47 & 63.34 &  {63.94} &  {63.44} &  {60.41} &  {60.51} &  {59.84} &  {60.55} \\ 
            3 & 57.56 & 55.90 & 57.96 & 57.96 & 63.78 & 61.93 &  {63.54} &  {59.49} &  {57.45} &  {56.34} &  {56.97} &  {60.30} \\ 
            4 & 57.44 & 59.83 & 57.37 & 58.09 & 64.21 & 62.94 &  {60.03} &  {62.23} &  {61.60} &  {56.69} &  {59.87} &  {62.26} \\ 
            5 & 59.49 & 58.57 & 57.41 & 59.30 &  {62.01} &  {60.54} &  {62.86} &  {62.13} &  {58.28} &  {57.48} &  {62.08} &  {57.26} \\ 
            Average & 57.56 & 58.88 & 57.68 & 59.15 & \textbf{63.76} & \textbf{62.44} & 62.48 & 61.52 & 60.56 & 58.64 & 59.8 & 60.23 \\
            $\sigma$ & 1.22 & 1.78 & 1.63 & \textbf{1.10} & \textbf{1.21} & 1.22 & 1.55 & 1.59 & 3.02 & 2.58 & 1.83 & 1.83\\ 
        \midrule
    \end{tabular}
    }
    \caption{Performance of the best individuals of the GBVSBP model for all \textbf{IMGSAL} database runs.}
    \label{tab:imgsal-results-gbvs-new}
\end{table}

Fitness results with the GBVSBP model and IMGSAL database show very similar behavior as we appreciate in outcomes. For example, the highest average was $63.76\%$ in the training stage with a 
$\sigma = 1.21$, while reaching a lowest average of $58.64\%$ with a $\sigma = 2.58$ in the testing stage. As we can appreciate, the results were not as high as the other two datasets since IMGSAL presents difficulties primarily due to poor segmentation when creating the ground truth. Also, the results took a long time to be completed (several months) due to the bigger image size. The algorithm reached the best solution on the third run-scoring $65.35\%$ in training and $63.45\%$ at testing. Table \ref{tab:imgsal-new-best-ind-estructure} provides the best solution.

\begin{table}[H]
    \centering
    \resizebox{\textwidth}{!}{
        \begin{tabular}{*{13}{c}}
            \hline
           \textbf{$EVO_{d}$ and $EFI$ operators} & \textbf{Fitness} \\
            \hline
            \multirow{4}{20cm}{$EVO_{O} = G_{(\sigma = 1)}(I_{v})$ \\ $EVO_{C} = \sqrt(I_{m})$ \\ $EVO_{S} = I_{m}$ \\ $EFI = |G_{(\sigma = 1)}(D_{y}(CM_{S}))|$} & Training = 0.6535 \\ & Testing = 0.6345 \\ & \\ & \\
            \hline
        \end{tabular}
    }
    \caption{Structure of the operators corresponding to the best solution for the \textbf{IMGSAL} database.}
    \label{tab:imgsal-new-best-ind-estructure}
\end{table}

\begin{table}[H]
    \centering
    \resizebox{\textwidth}{!}{
        \begin{tabular}{*{13}{c}}
        \toprule
        \multicolumn{13}{c}{\textbf{PASCAL-S}} \\
        Fold & \multicolumn{2}{c}{Run 1} & \multicolumn{2}{c}{Run 2} & \multicolumn{2}{c}{Run 3} & \multicolumn{2}{c}{Run 4} & \multicolumn{2}{c}{Run 5} & \multicolumn{2}{c}{Run 6} \\ 
        \midrule
         & Trng & Test & Trng & Test & Trng & Test & Trng & Test & Trng & Test & Trng & Test\\ 
            1 & 62.03 & 59.75 & 61.39 & 57.53 & 61.88 & 57.98 & 60.76 & 56.06 & 61.62 & 58.75 & 60.89 & 58.19 \\ 
            2 & 61.53 & 55.12 & 61.95 & 54.99 & 61.11 & 56.52& 61.33 & 52.61 & 61.23 & 53.64 & 61.01 & 56.70\\ 
            3 & 59.28 & 58.83 & 60.08 & 61.08 & 59.84 & 60.32 & 60.60 & 59.57 & 59.82 & 61.33 & 59.71 & 61.78 \\ 
            4 & 59.57 & 58.36 & 58.20 & 54.28 & 58.59 & 55.91 & 58.61 & 58.90 & 58.21 & 57.13 & 57.89 & 56.44 \\ 
            5 & 63.20 & 61.40 & 61.94 & \textbf{61.94} & 63.08 & 59.12 & \textbf{63.38} & 59.01 & 62.93 & 59.00 & 62.87 & 61.34 \\ 
            Average & \textbf{61.12} & 58.69 & 60.71 & 57.96 & 60.90 & 57.97 & 60.94 & 57.23 & 60.76 & 57.97 & 60.47 & \textbf{58.87}  \\
            $\sigma$ & 1.67 & 2.31 & \textbf{1.60} & 3.47 & 1.75 & \textbf{1.82} & 1.71 & 2.92 & 1.81 & 2.85 & 1.83 & 2.54\\ 
        \midrule
    \end{tabular}
    }
    \caption{Performance of the best individuals of the GBVSBP model for all \textbf{PASCAL-S} database runs.}
    \label{tab:pascal-results-gbvs-new}
\end{table}

The experimental results with GBVSBP for the PASCAL-S database have higher similarity because the F-measure has greater stability than previous results. The results show a stable standard deviation maintained between  1.60 and 3.47 for training and testing. We obtain similar behavior in the average results between $60.47\%$ and $61.12\%$ for the training stage. The best solution was reached in the fifth fold at the fourth run-scoring $63.38\%$ during training, while in testing, the algorithm scored $61.94\%$. Table \ref{tab:pascal-new-best-ind-estructure} shows the best trees.

\begin{table}[H]
    \centering
    \resizebox{\textwidth}{!}{
        \begin{tabular}{*{13}{c}}
            \hline
           \textbf{$EVO_{d}$ and $EFI$ operators} & \textbf{Fitness} \\
            \hline
            \multirow{4}{20cm}{$EVO_{O} = G_{(\sigma = 1)}(G_{(\sigma = 1)}(I_{v}))$ \\ $EVO_{C} = I_{m}$ \\ $EVO_{S} = \lfloor(dilation_{disk}(dilation_{square}(erosion_{disk}(I_{b}))))\rfloor$ \\ $EFI = |G_{(\sigma = 2)}(D_{y}(CM_{MM}))|$} & Training = 0.6338 \\ & Testing = 0.6194 \\ & \\ & \\
            \hline
        \end{tabular}
    }
    \caption{Structure of the operators corresponding to the best solution for the \textbf{PASCAL-S} database.}
    \label{tab:pascal-new-best-ind-estructure}
\end{table}

\begin{table}[H]
    \centering
    \resizebox{\textwidth}{!}{
        \begin{tabular}{*{13}{c}}
            \toprule
            \multicolumn{13}{c}{\textbf{FT}} \\
            Fold & \multicolumn{2}{c}{Run 1} & \multicolumn{2}{c}{Run 2} & \multicolumn{2}{c}{Run 3} & \multicolumn{2}{c}{Run 4} & \multicolumn{2}{c}{Run 5} & \multicolumn{2}{c}{Run 6} \\ 
            \midrule
             & Trng & Test & Trng & Test & Trng & Test & Trng & Test & Trng & Test & Trng & Test\\ 
                1 & 88.41 & 88.51 & 88.07 & 86.00 & 87.92 & 80.83 & 88.32 & 85.75 & 87.52 & \textbf{89.02} & 87.04 & 86.69\\ 
                & \multicolumn{2}{c}{Run 1} & \multicolumn{2}{c}{Average} & \multicolumn{2}{c}{$\sigma$} & \multicolumn{2}{c}{Minimum} & \multicolumn{2}{c}{Maximum} & \multicolumn{2}{c}{Mean} \\
                & Trng & Test & Trng & Test &  Trng & Test & Trng & Test &  Trng & Test  &  Trng & Test  \\ 
              2  & \textbf{89.02} & 88.56 & 88.24 & 86.53 & \textbf{0.56} & 3.75 & 87.04 & 80.83 & 89.09 & 89.02 & 88.07 & 86.69 \\ 
            \midrule
        \end{tabular}
    }
    \caption{Performance of the best individuals of the GBVSBP + MCG model for all \textbf{FT} database runs.}
    \label{tab:ft-results-gbvs-mcg-new}
\end{table}

The experimental results with GBVSBP + MCG for the FT database show some loss of aptitude against its counterpart in the first block of experiments ($5\%$ in training and $3\%$ in tests). However, the performance values are more stable with $\sigma = 0.56$ for training and $\sigma = 0.49$ for testing. In addition to stability, we must bear in mind that the results of this experiment will be much more consistent when the best individual is tested in the benchmark since we are using the same F-measure. As a result, the best results were $89.02\%$ in training corresponding to the second run of the first fold and $89.02\%$ at testing discovered in the first run of the fifth fold. Table \ref{tab:ft-gbvs-mcg-new-best-ind-estructure} gives the best set of trees.

\begin{table}[H]
    \centering
    \resizebox{\textwidth}{!}{
        \begin{tabular}{*{13}{c}}
            \hline
           \textbf{$EVO_{d}$ and $EFI$ operators} & \textbf{Fitness} \\
            \hline
            \multirow{4}{20cm}{$EVO_{O} = G_{(\sigma = 1)}(D_{y}(I_{y}))$ \\ $EVO_{C} = ((DKL_{\Phi}^{1/0.62}) - (((Exp(I_{b})^{1/0.62})^{1/0.62}) - ((DKL_{\Phi} - DKL_{\Phi}) - (I_{b} - I_{b}))))$ \\ $EVO_{S} = I_{y}$ \\ $EFI = CM_{MM}$} & Training = 0.8902 \\ & Testing = 0.8902 \\ & \\ & \\ & \\
            \hline
        \end{tabular}
    }
    \caption{Structure of the operators corresponding to the best solution for the \textbf{FT} database. Note that we can simplify the second tree. However, we report the programs as returned by the computer.}
    \label{tab:ft-gbvs-mcg-new-best-ind-estructure}
\end{table}

\subsection{Analysis of the Best Evolutionary Run}

\begin{figure}
\resizebox{\textwidth}{!}
{
    \centering
\begin{tabular}{cc}
    \includegraphics[scale=0.3]{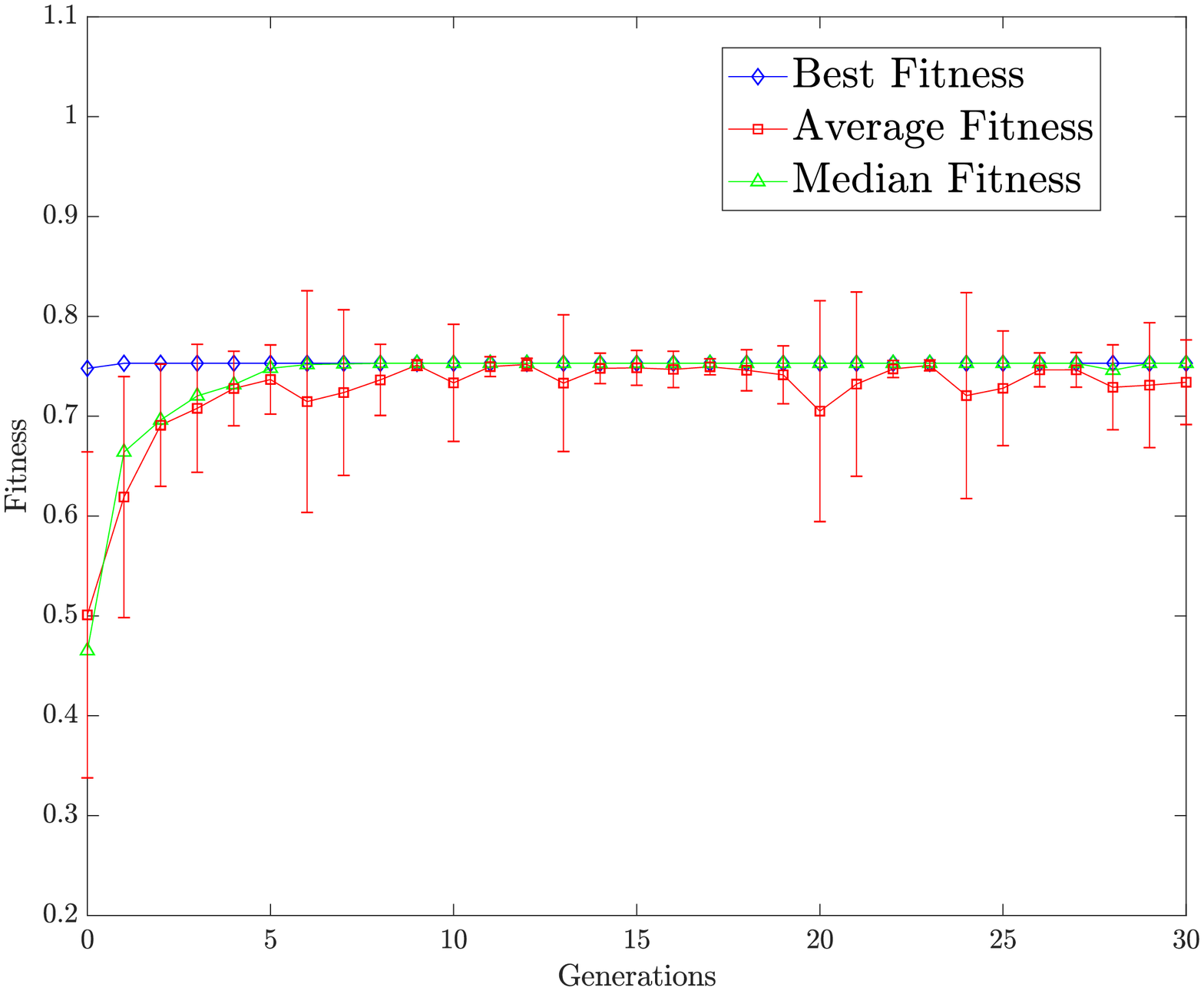} &
    \includegraphics[scale=0.3]{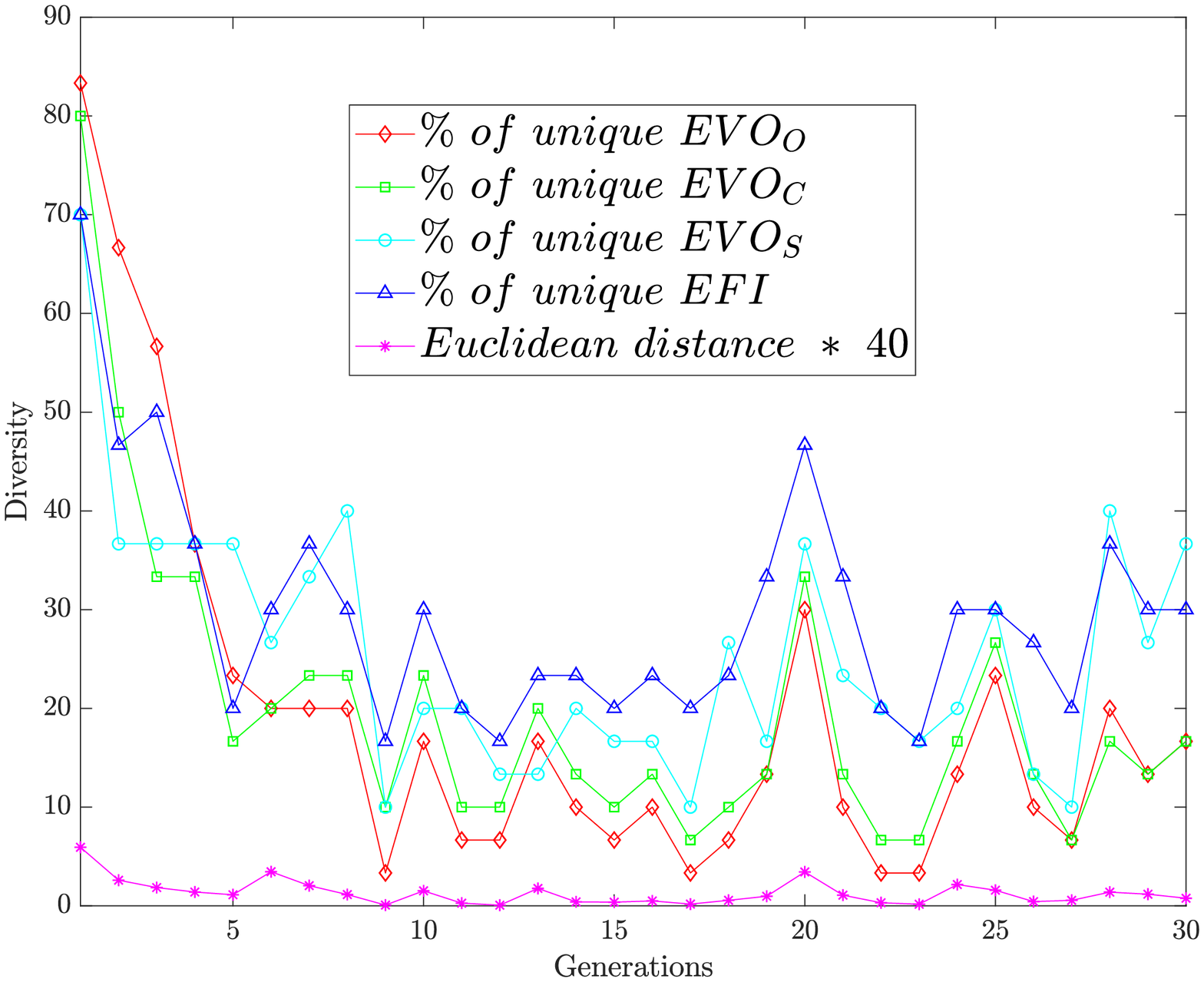} \\
    \includegraphics[scale=0.3]{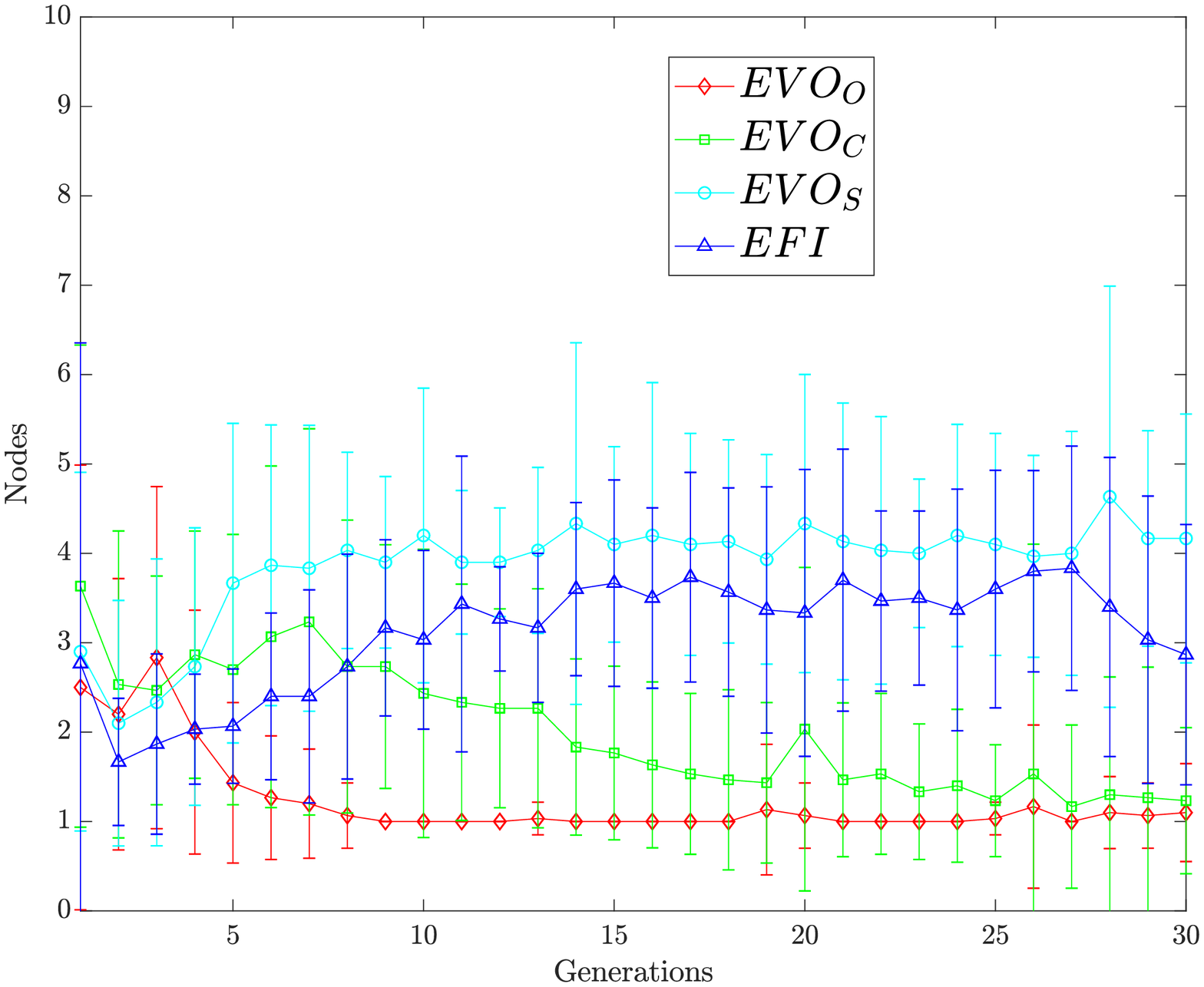} &
    \includegraphics[scale=0.3]{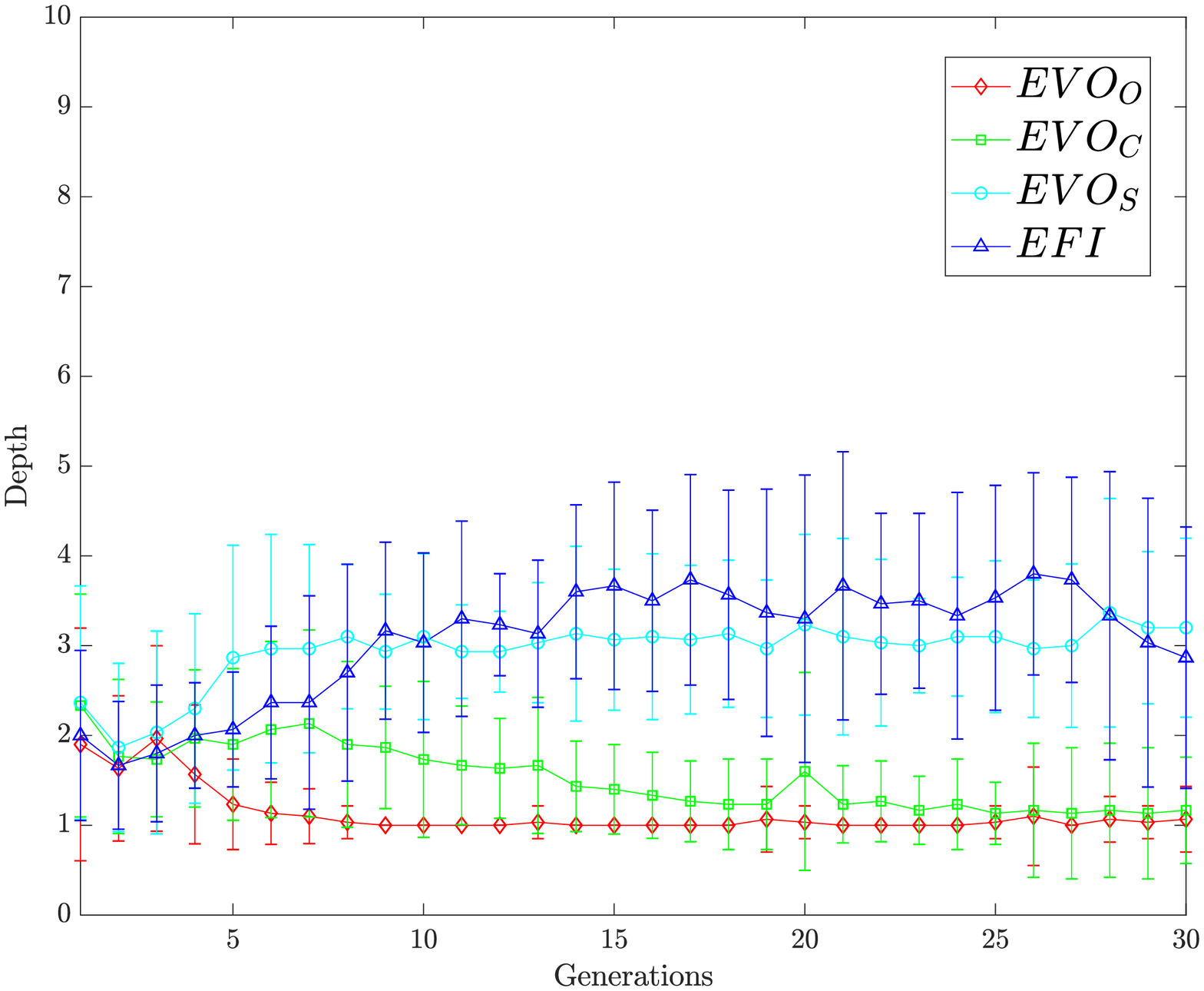} \\
\end{tabular}
}
    \caption{Brain programming statistics of the run corresponding to the best GBVSBP model for the FT database.}

    \label{fig:ft-gbvs-ind-results}

\end{figure}

Typical experimental results that illustrate the inner workings of genetic programming are those related to fitness, diversity, number of nodes, and depth of the tree.
Figure \ref{fig:ft-gbvs-ind-results} provides charts giving best fitness, average fitness, and median fitness. The purpose is to detail the performance and complexity of solutions through the whole evolutionary run. As we can observe, artificial evolution scores a high fitness within the first generations. On average, BP converges around the seventh generation. The chart depicting diversity shows the convergence of solutions in all of the four trees characterizing the program. Compared to the fitness plot, these data demonstrate that despite the differences in diversity that occurred during the experiment, the model's performance remained constant. One of the biggest problems using genetic programming is incrementing a program's size without a rise in the program's performance, mainly when the final result cannot generalize to new data. This problem is called bloat and is usually associated with tree representation. As observed in the last two graphs, the complexity is kept low with the number of nodes below seven and depth below five regarding all trees. These numbers were consistently below the proposed setup for all experiments. The hierarchical structure allows an improvement in performance and the management of the algorithm's complexity.

\subsection{Comparison with Other Approaches}

\begin{table}[H]

    \centering

    \begin{tabular}{*{2}{c}}

        \toprule

        \textbf{Saliency Model} & \textbf{Score (F-measure)} \\ 

        \midrule
        
        GBVSBP + MCG & \textbf{86.72} \\ 
        
        SF (\cite{Krahenbuhl2012}) & 85.38 \\ 
        
        GBVS + MCG (\cite{Li2014}) & 85.33 \\ 
        
        PCAS (\cite{Margolin}) & 83.93 \\ 

		GC (\cite{Cheng2011}) & 80.64 \\ 

        DHSNET (\cite{Contreras-Cruz2019}) & 74.06 \\ 
		
		FT (\cite{Achantay2009}) & 71.23 \\        
		
		GBVSBP & 69.08 \\ 
        
        GBVS (\cite{Li2014}) & 65.25 \\

        FOA (\cite{Dozal2014}) & 60.05 \\ 
        \toprule

    \end{tabular}

    \caption{Comparison using the benchmark with other algorithms in the FT database (\cite{Li2014}).}

    \label{tab:results-gbvs-mcg-vs-cnn}

\end{table}

\begin{figure*}

\resizebox{\textwidth}{!}
{
    \centering

    \includegraphics[scale=0.15]{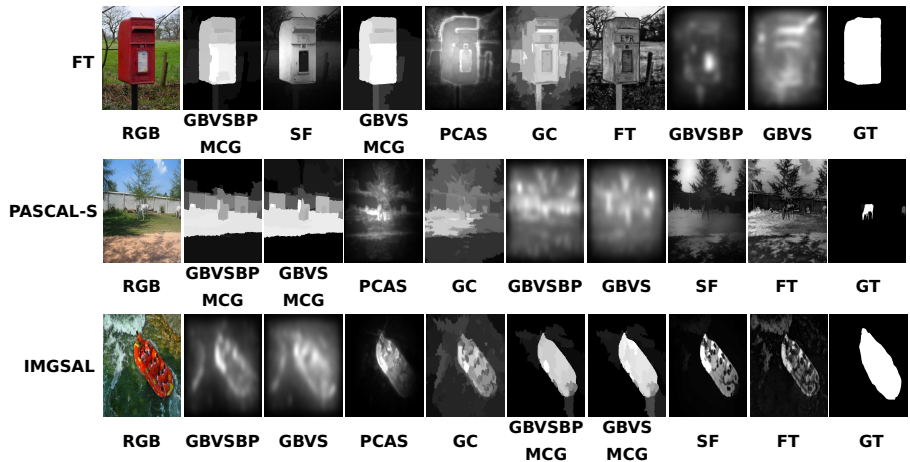}
}
    \caption{Example of outstanding maps obtained in experiments where our models observe better performance.}

    \label{fig:image_results}

\end{figure*}

\begin{table*}
    \centering
       \resizebox{\textwidth}{!} 
{
        \begin{tabular}{*{8}{c}}
            \toprule
            \textbf{Dataset} & \textbf{FT} & \textbf{HC} & \textbf{MDC} & \textbf{MBS} & \textbf{MIN} & \textbf{MAX} & \textbf{AVG} \\ 
            \midrule 
            MSRA-BTest & 0.5319 & 0.5663 & 0.7241 & 0.7116 & 0.6762 & 0.6105 & 0.6751 \\
            
            ECSSD & 0.3775 & 0.3894 & 0.6127 & 0.5570 & 0.5177 & 0.4344 & 0.4994 \\
            
            SED2 & 0.6250 & 0.6079 & 0.6074 & 0.6654 & 0.6833 & 0.6106 & 0.6971 \\
            
            iCoseg & 0.5545 & 0.5471 & 0.6348 & 0.6253 & 0.6210 & 0.5587 & 0.6105 \\
            
            \textbf{Dataset} & \textbf{TOP2} & \textbf{CPSO} & \textbf{GA} & \textbf{PSO} & \textbf{GPMCC} & \textbf{GPSED} & \textbf{GBVSBP+MCG} \\
            
            MSRA-BTest & 0.7423 & 0.7137 & 0.7579 & 0.7455 & 0.7711 & 0.7662 & \textbf{0.8308} \\
            
            ECSSD & 0.5979 & 0.5588 & 0.6200 & 0.5988 & 0.6592 & 0.6592 & \textbf{0.7591} \\
            
            SED2 & 0.6658 & 0.6653 & 0.6914 & 0.6701 & 0.7148 & \textbf{0.7340} & 0.6919 \\
            
            iCoseg & 0.6556 & 0.6235 & 0.6678 & 0.6557 & 0.6865 & 0.7157 & \textbf{0.7168} \\
            \toprule
        \end{tabular}
    \caption{Comparison with others models and databases published 
    in \cite{Contreras-Cruz2019}.}
    \label{tab:results-gbvs-mcg-vs-contreras}
}
\end{table*}

\begin{figure*}

    \centering
    
    \includegraphics[width=0.7\textwidth]{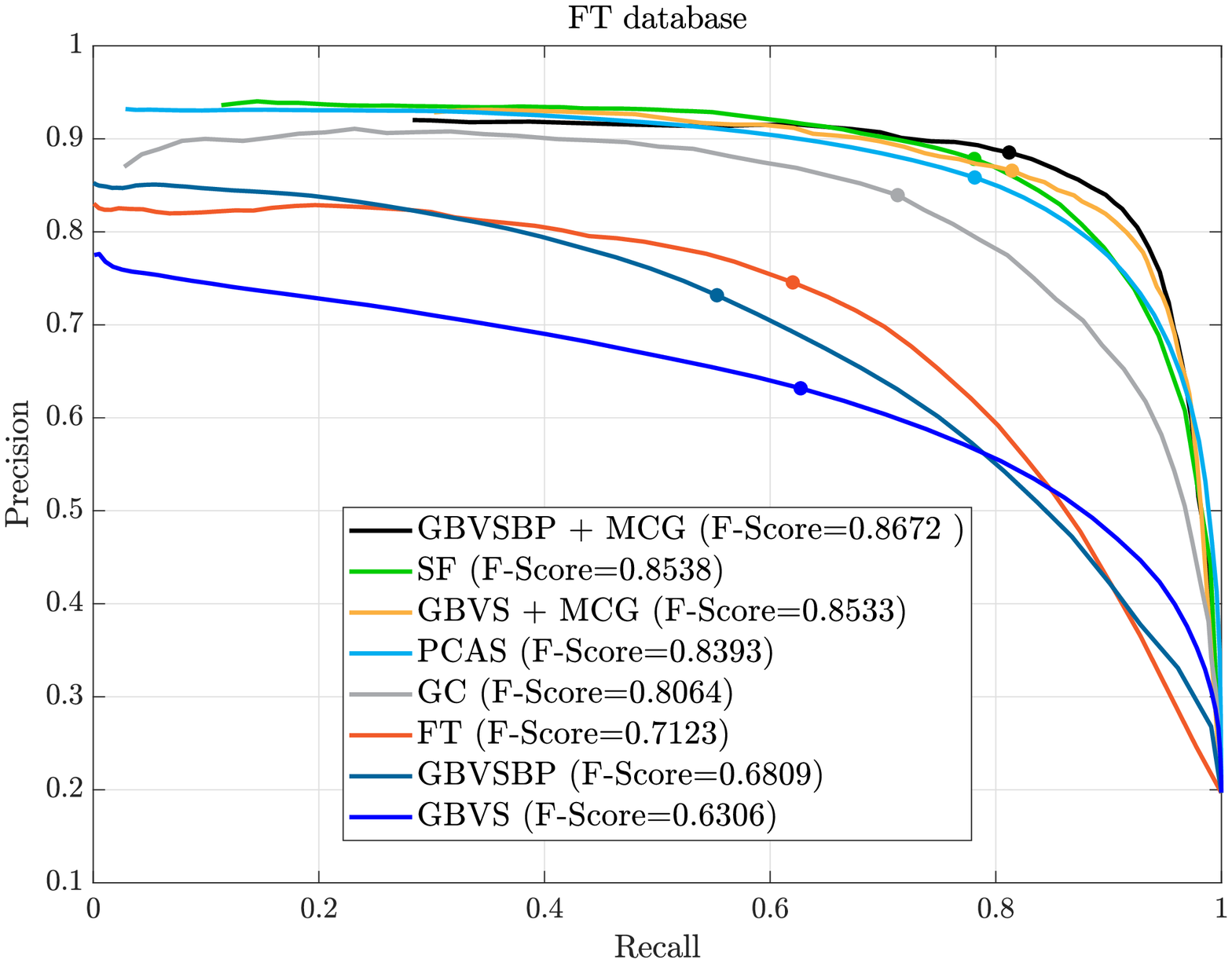}
    \includegraphics[width=0.7\textwidth]{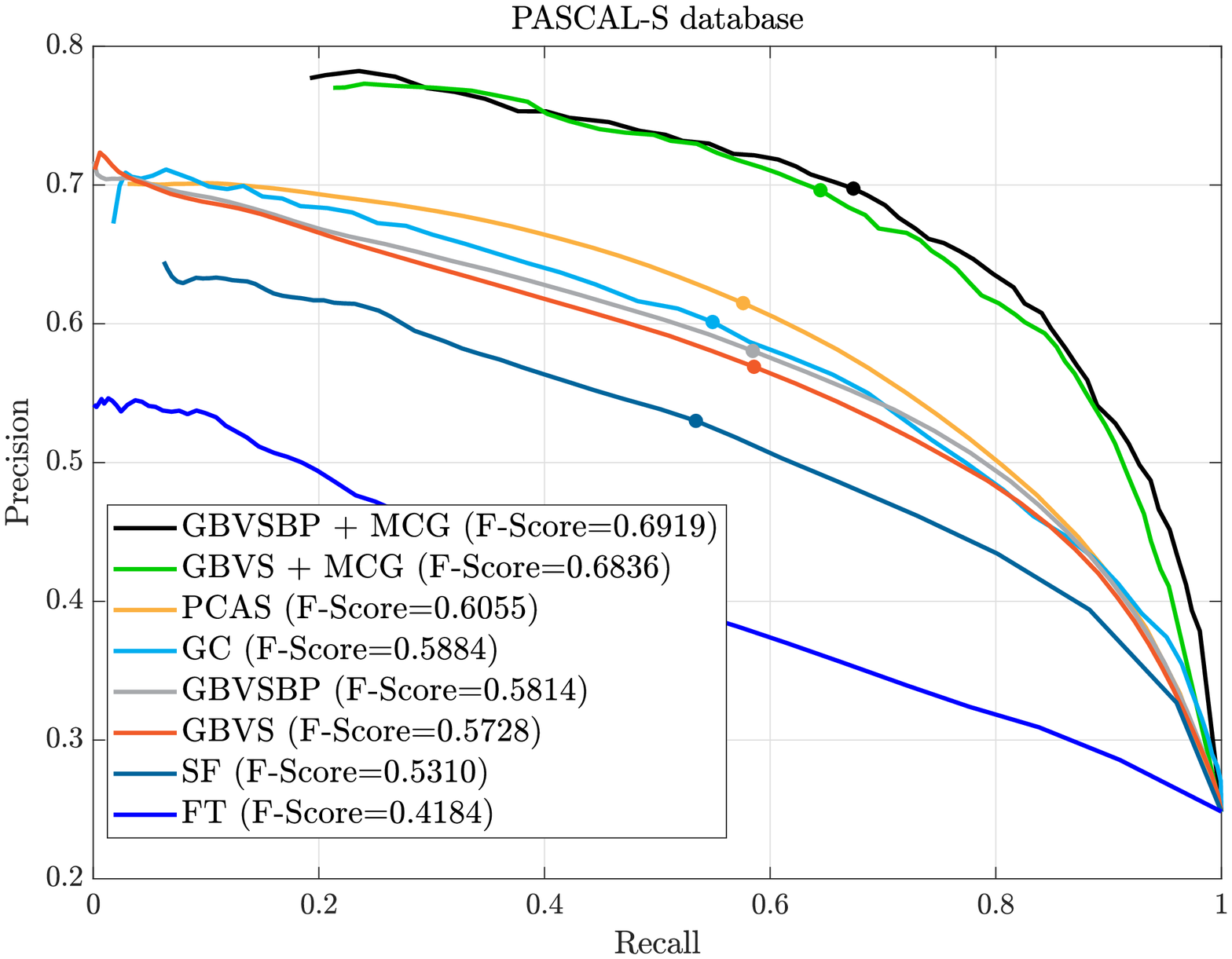}
    \includegraphics[width=0.7\textwidth]{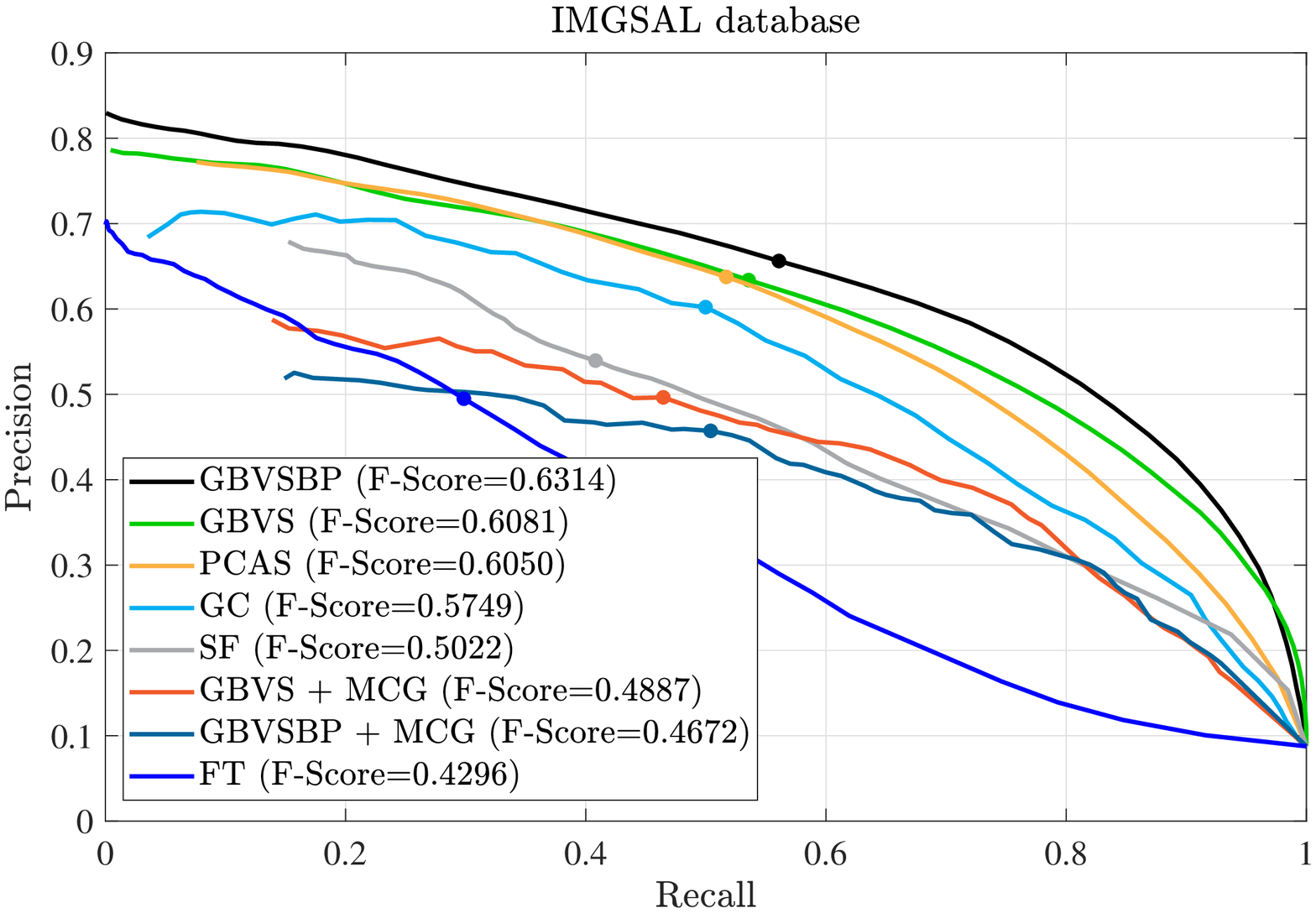}

    \caption{The precision-Recall curve for FT, PASCAL-S, and IMGSAL datasets using the following algorithms: GBVSBP + MCG, SF, GBVS + MCG, PCAS, GC, FT, GBVSBP, and GBVS.}

    \label{fig:gpft-results}

\end{figure*}

The values presented above correspond to fitness in the evolutionary cycle of BP. The benchmark offers two modalities: one which uses only $60\%$ of the database, and another containing all images; we keep the first option. Table \ref{tab:results-gbvs-mcg-vs-cnn} shows final results achieved on the testing set over 10 random splits with our best program considering the FT dataset. Here, we appreciate final results considering the following algorithms for salient object detection: FT--Frequency-tuned, GC--Global Contrast, SF--Saliency Filters, PCAS--Principal Component Analysis, and DHSNET. Also, we include the original proposal of the artificial dorsal stream named focus of attention (FOA) reported in \cite{Dozal2014}. Note that we overpass all other algorithms in the benchmark. Figure \ref{fig:image_results} presents image results of all algorithms in the three databases for visual comparison. Figure \ref{fig:gpft-results} provides Precision-Recall curves for the FT, IMGSAL, and PASCAL-S databases of the benchmark. Again we score highest in the IMGSAL and PASCAL-S datasets. Note that even if the computer model is symbolic, the interpretation remains numeric, and therefore exists computer errors. Anyway, the computation is data-independent since the proposal follows a function-driven paradigm. 

\begin{figure}
    \centering
    \includegraphics[scale=0.25]{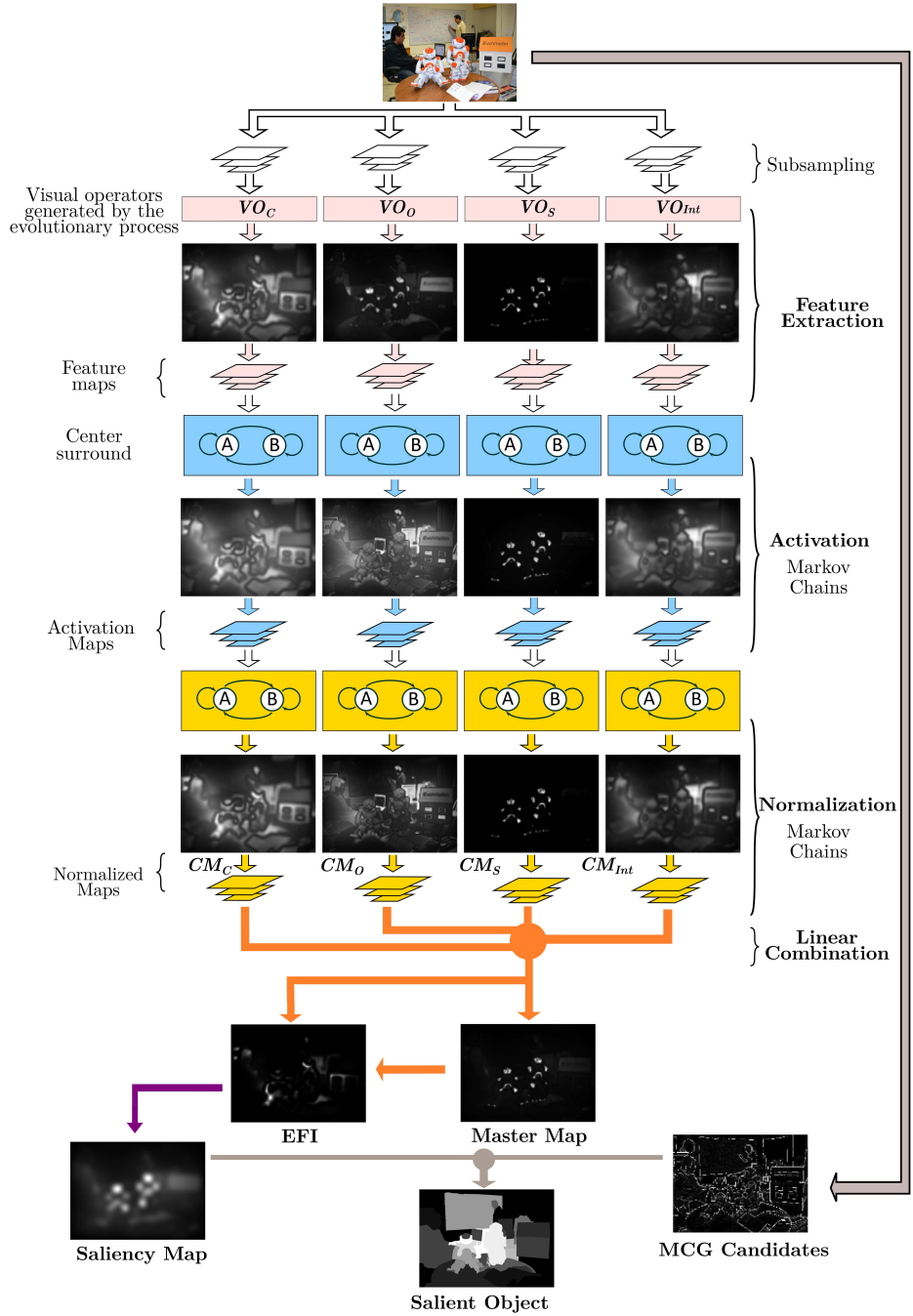}
    \caption{Brain programming result of the dorsal stream using the best GBVS+MCG program.}
    \label{fig:evoGBVS-image}
\end{figure}

Finally, we test our best solution (GBVSBP + MCG) on four databases studied in \cite{Contreras-Cruz2019}, and the results are in Table \ref{tab:results-gbvs-mcg-vs-contreras}. The best solution designed by {\it Contreras-Cruz et al.} was trained with MSRA-A and then it was tested with MSRA-BTest, ECSSD, SED2, and iCoseg. We observe that we score highest on three datasets MSRA-BTest, ECSSD, and iCoseg while achieving competitive results on SED2. We provide such comparison since \cite{Contreras-Cruz2019} does not test their algorithms with the benchmark protocol. Therefore it is hard to make a clear comparison between both approaches, and the results we provide here serve the purpose of illustrating the methodologies' performance. Figure \ref{fig:evoGBVS-image} illustrates the image processing through the whole GBVSBP+MCG program.

\section{Conclusions}


In this work, we propose a method to improve the ADS model presented by \cite{Dozal2014}. This method consists of applying an algorithm called GBVS that surpasses Itti's previous model. GBVS uses a graph-based approach using Markov chains while involving the same stages as the Itti model. Moreover, we follow the idea of combining fixation prediction with a segmentation algorithm to obtain a new method called GBVSBP + MCG to tackle the problem of salient object segmentation. As we show in the experiments, the novel design scores highest in the FT, IMGSAL, and PASCAL-S datasets of a benchmark provided by \citep{Li2014}. These tests show the strength and generalization power of the discovered model compared to others developed manually and current CNNs such as DHSNET, surpassed by more than 12 percentage points in FT. Also, we give results on four datasets described in \cite{Contreras-Cruz2019} with outstanding results. The results are revealing about the difficulty of solving this visual task. In the FT and PASCAL-S, objects are defined with accurate ground truth, and the algorithm GBVSBP + MCG improves the results of the original algorithm slightly. However, in the IMGSAL database, the ground truth is poorly segmented, and GBVSBP significantly improves the score compared to the original proposal. Therefore, we can say that there is a considerable benefit in combining analytical methods with heuristic approaches. We believe that this mixture of strategies can help find solutions to challenging problems in visual computing and beyond. One advantage is that the overall process and final designs are explainable, which is considered a hot topic in today's artificial intelligence. This research attempts to advance studies conducted by experts (neuroscientists, psychologists, and computer scientists) by adapting the symbolic paradigm for machine learning to find better ways of describing the brain's inner workings.

\bibliography{mybibfile}

\end{document}